%% file: main.tex
\documentclass[10pt,twocolumn,letterpaper]{article}

\usepackage[pagenumbers]{cvpr} 

\input{preamble}

\definecolor{cvprblue}{rgb}{0.21,0.49,0.74}
\usepackage[pagebackref,breaklinks,colorlinks,allcolors=cvprblue]{hyperref}
\usepackage{xcolor} 

\def\paperID{10570} 
\def\confName{CVPR}
\def\confYear{2025}

\title{Symmetry Strikes Back: From Single-Image Symmetry Detection \\ to 3D Generation}

\author{Xiang Li$^{1}$\hspace{0.23in}Zixuan Huang$^{1}$\hspace{0.23in} Anh Thai$^{{2}}$\hspace{0.23in} James M. Rehg$^{{1}}$\\
$^1$University of Illinois at Urbana-Champaign\quad \\ 
$^2$Georgia Institute of Technology\\
}

\begin{document}

\twocolumn[{%
\renewcommand\twocolumn[1][]{#1}%
\maketitle
\vspace{-6ex}
\begin{center}
\textbf{\url{https://ryanxli.github.io/reflect3d}}
\end{center}%
\begin{center}
    \centering
    \captionsetup{type=figure}
    \includegraphics[width=\textwidth,trim={1cm 4.5cm 1.7cm 5cm}, clip]{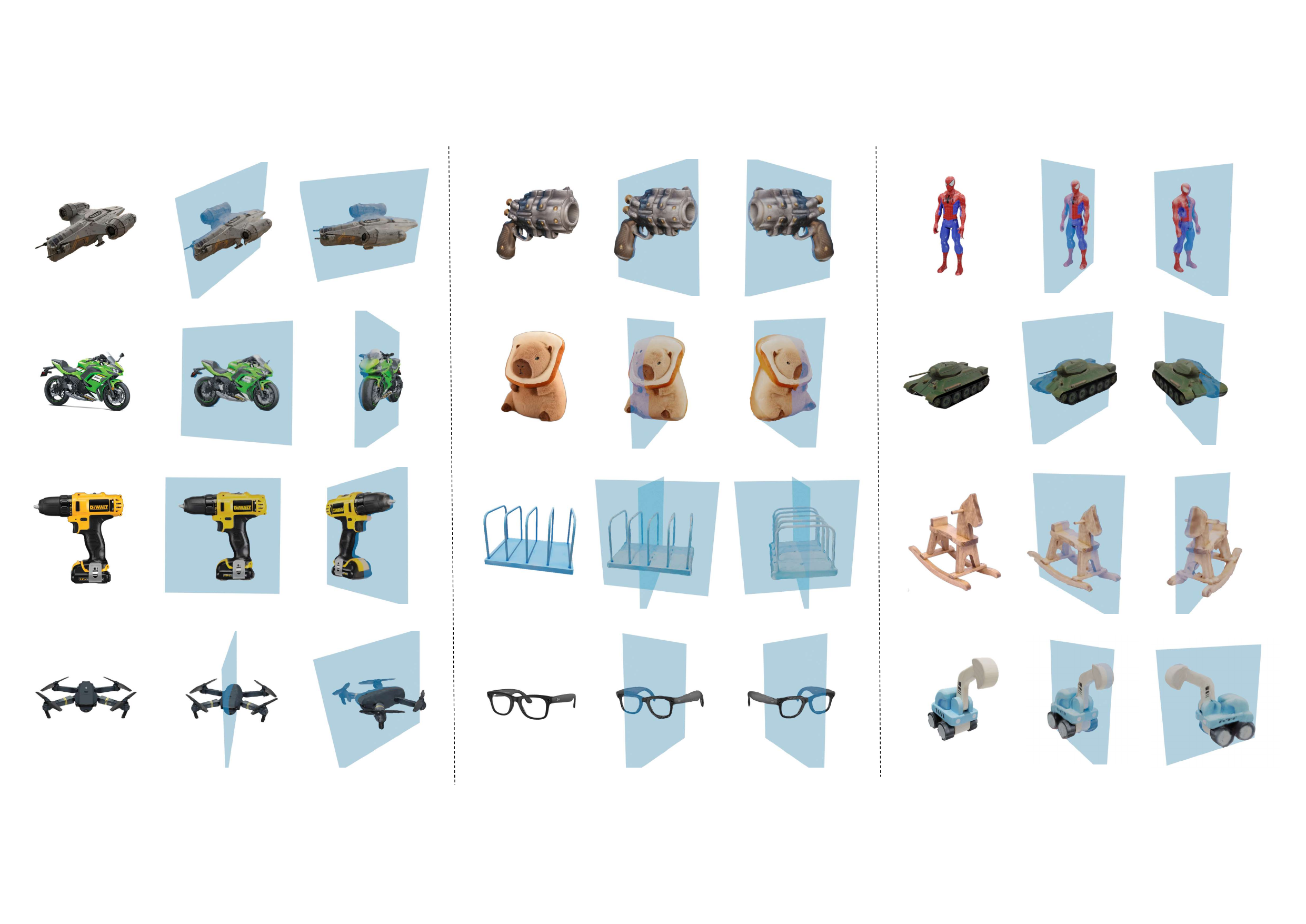}
    \vspace{-5ex}
    \captionof{figure}{We propose~\methodname, a zero-shot 3D reflection symmetry detector capable of accurately detecting 3D symmetry from a single RGB image of an arbitrary object. Conditioned on the detected symmetry, we improve single-image 3D generation in both geometry and texture quality.  }
\end{center}%
}]



\input{sec/0_abstract}    
\input{sec/1_intro}

\input{sec/2_related}

\input{sec/3_method}

\input{sec/4_data}

\input{sec/5_experiments}
\input{sec/6_conclusion}
\newpage
\newpage
\newpage
\newpage
{
    \small
    \bibliographystyle{ieeenat_fullname}
    \bibliography{main}
}

\input{sec/X_suppl}

\end{document}

%% file: preamble.tex
\def\methodname{Reflect3D\xspace}


%% file: sec/0_abstract.tex
\begin{abstract}

\vspace{-1ex}


Symmetry is a ubiquitous and fundamental property in the visual world, serving as a critical cue for perception and structure interpretation. This paper investigates the detection of 3D reflection symmetry from a single RGB image, and reveals its significant benefit on single-image 3D generation. 
We introduce \methodname, a scalable, zero-shot symmetry detector capable of robust generalization to diverse and real-world scenarios. Inspired by the success of foundation models, our method scales up symmetry detection with a transformer-based architecture. We also leverage generative priors from multi-view diffusion models to address the inherent ambiguity in single-view symmetry detection. Extensive evaluations on various data sources demonstrate that \methodname establishes a new state-of-the-art in single-image symmetry detection. Furthermore, we show the practical benefit of incorporating detected symmetry into single-image 3D generation pipelines through a symmetry-aware optimization process. The integration of symmetry significantly enhances the structural accuracy, cohesiveness, and visual fidelity of the reconstructed 3D geometry and textures, advancing the capabilities of 3D content creation.

\end{abstract}


%% file: sec/1_intro.tex
\vspace{-3ex}
\section{Introduction}

"Symmetry is what we see at a glance." — Blaise Pascal. \\

Symmetry is ubiquitous in both natural and artificial objects. It serves as a crucial visual cue that helps humans perceive object structure and interpret spatial relationships~\cite{treder2010behind}. 
In computer vision, symmetry has long been leveraged as a structural constraint to simplify complex visual tasks, such as pose estimation~\cite{merrill2022symmetry, zhao2023learning}, grasp detection~\cite{shi2022symmetrygrasp}, and 3D reconstruction~\cite{wu2020unsup3d, xu2020ladybird, wu2021derender, yao2020front2back}. While existing methods achieve impressive symmetry detection results from 3D or depth data~\cite{Mitra2006PartialAA, ji2019fast, gao2020prs, shi_siga20, shi2022learning, zhang2023single, je2024robustlangsym}, detecting symmetry from a single RGB image remains challenging and underexplored. Past approaches have made progress with restricted object categories~\cite{zhou2021nerd, lin2021nerd++}. However, robust zero-shot symmetry detection from a single image of arbitrary objects remains an open challenge.

Inspired by the transformative impact of foundation models in NLP~\cite{gpt32020}, vision~\cite{dinov2, sam2023}, and speech~\cite{whisper2023}, we revisit single-image symmetry detection through a foundation model perspective. The key to the success of foundation models has been the combination of versatile transformer-based architectures and large, diverse datasets. Based on this observation, we design a transformer-based model for symmetry detection with minimal use of explicit 3D inductive priors. We train this model end-to-end on our large-scale 3D symmetry dataset that combines Objaverse~\cite{objaverse} and ShapeNet~\cite{chang2015shapenet}.

One of the major challenges in single-image symmetry detection is single-view ambiguity~\cite{zhou2021nerd}, where depth, perspective distortion, and occlusion obscure symmetry cues. We propose to address this problem by leveraging a generative prior. We apply multi-view diffusion models~\cite{liu2023zero1to3} to synthesize surrounding views of the target object, providing a more complete and less ambiguous perspective on symmetry. By aggregating symmetry predictions across multiple views, we achieve comprehensive and accurate symmetry detection.


Specifically, our approach introduces a novel pipeline for single-image symmetry detection. We first generate multiple surrounding views of the object using a multi-view diffusion model. Next, we apply our feed-forward symmetry detector across each view. Our symmetry detector includes three modules: 1) a frozen DINOv2 encoder that extracts geometry-aware image features, 2) a transformer-based symmetry decoder that uses cross-attention to query these features with various symmetry hypotheses, and 3) MLP heads that classify these hypotheses and refine them through a regressive correction stage. Finally, predictions from all views are clustered and aggregated, yielding precise and comprehensive symmetry predictions.

We evaluate our detected symmetry on two challenging real-world scanned-object datasets, GSO~\cite{gso} and OmniObject3D~\cite{omniobj}. Despite the significant domain shifts from synthetic training data, our model demonstrates remarkable generalization ability. Even without multi-view aggregation, our feed-forward symmetry detector already achieves state-of-the-art performance. With multi-view aggregation, our symmetry detection performance further improves by a significant margin. We also illustrate the robustness of our approach on in-the-wild internet images.

Finally, we demonstrate the practical utility of symmetry detection in improving single-image 3D generation pipelines. By integrating symmetry predictions into the Score Distillation Sampling (SDS) optimization process, we observe improved 2D and 3D fidelity across both real-world datasets and in-the-wild images, highlighting the broader impact of our symmetry detection approach.


%% file: sec/2_related.tex
\section{Related Work}

\begin{figure*}[t]
  \centering
  \includegraphics[width=\textwidth,trim={2cm 5.5cm 2cm 5cm}, clip]{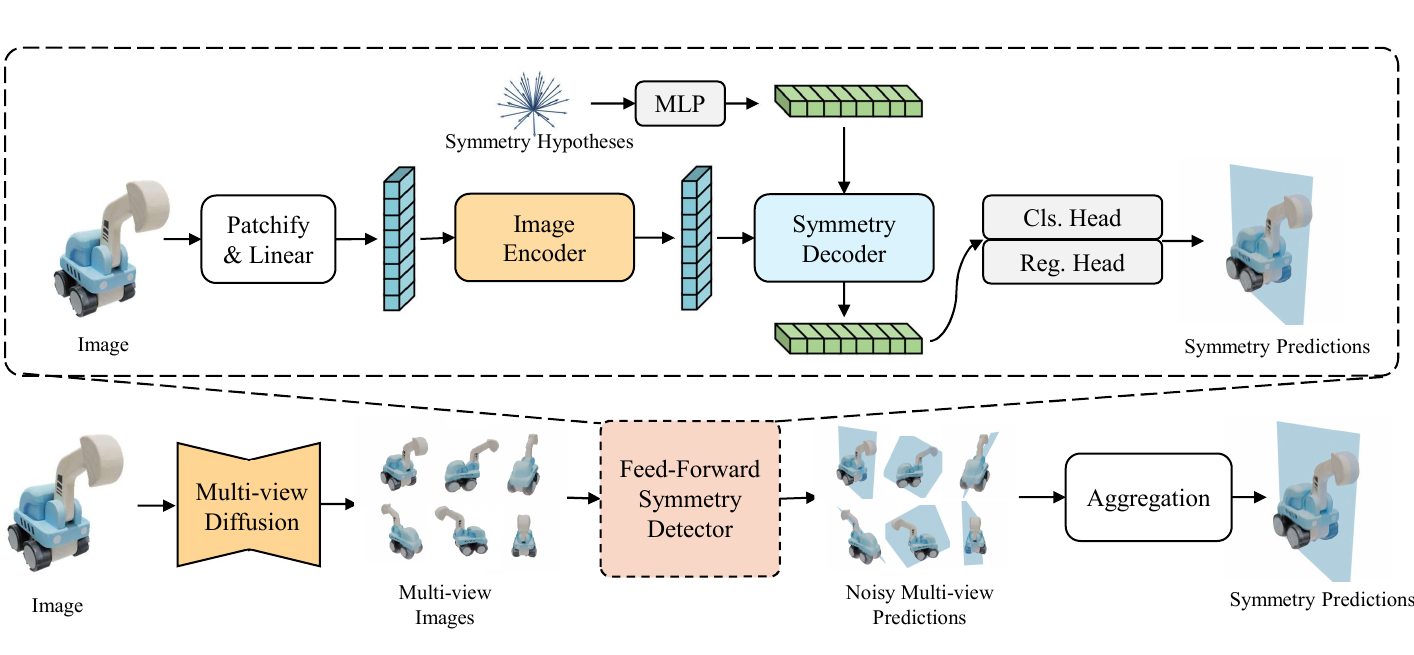}
  \vspace{-2.5em}
   \caption{Overview of~\methodname, our zero-shot single-image symmetry detector. Top: Our transformer-based feed-forward symmetry detector (\Cref{sec:ff}) predicts symmetry planes from a single RGB image. Bottom: Our multi-view symmetry enhancement pipeline (\Cref{sec:mv}) leverages multi-view diffusion to resolve the inherent single-view ambiguity in symmetry detection. Aggregating symmetry predictions from multiple synthesized views results in more precise and comprehensive symmetry predictions. }
   \vspace{-1ex}
   \label{fig:symm_method}
\end{figure*}

\noindent\textbf{Symmetry Detection.} Early works on symmetry detection introduce several foundational algorithms~\cite{atallah1985symmetry, wolter1985optimal, zabrodsky1995symmetry, kiryati1998detecting}. Most of these efforts focus on detecting 3D symmetry from explicit 3D representations. For instance, ~\cite{Mitra2006PartialAA} introduces methods for detecting partial and approximate 3D symmetry. More recent works~\cite{ji2019fast, gao2020prs, e3sym} apply neural networks to the problem. Specifically,~\cite{je2024robustlangsym} leverages diffusion models to detect partial symmetry in noisy 3D data, achieving strong performance but still relying on 3D input. Several papers~\cite{shi_siga20, shi2022learning, shi2022symmetrygrasp, zhang2023single} extend 3D symmetry detection to RGB-D or depth data. They unproject depth into incomplete 3D point clouds and detect symmetry from the 3D point clouds. Despite these efforts, detecting 3D symmetry directly from 2D RGB image remains challenging and underexplored. \cite{zhou2021nerd} introduces NeRD, a neural reflection symmetry detector that constructs 3D cost volumes from RGB image and camera parameters, then detects symmetry based on these cost volumes. 
While NeRD~\cite{zhou2021nerd} and its successor NeRD++~\cite{lin2021nerd++} achieve reasonable results on in-domain categories and images, they struggle to generalize to in-the-wild scenarios. In contrast, our \methodname demonstrates a strong zero-shot generalization ability to diverse real-world imagery.

\noindent\textbf{Single-image to 3D.} 
Single-image 3D reconstruction or generation aims to infer 3D structures from a single 2D image. This is a highly challenging task due to the inherent ambiguity of inferring 3D from only a single viewpoint. Recent work has made significant progress and can be classified into three main directions: 1) feed-forward 3D reconstruction~\cite{huang2024zeroshape, hong2024lrm, tang2024lgm, wang2024crm}, 2) diffusion-based 3D generation~\cite{jun2023shap, lan2024ln3diff}, and 3) optimization-based approaches~\cite{poole2022dreamfusion, tang2023dreamgaussian}. In our work, we take optimization-based approaches as a case study, where we integrate our detected symmetry into an optimization pipeline. Specifically, most of the recent optimization-based 3D generation methods are based on Score Distillation Sampling (SDS). DreamFusion~\cite{poole2022dreamfusion} initially proposes SDS. The high-level idea is to leverage a pre-trained 2D diffusion model as an image prior to regularize the optimization of a Neural Radiance Field (NeRF)~\cite{mildenhall2020nerf}. Zero-1-to-3~\cite{liu2023zero1to3} introduces an image-conditioned novel view synthesis diffusion model, which can be used to improve SDS guidance. DreamGaussian~\cite{tang2023dreamgaussian} incorporates the 3D Gaussian splatting~\cite{kerbl3Dgaussians} representation to improve the efficiency and quality of SDS-guided 3D generation. Our approach to leveraging symmetry detection for reconstruction builds on DreamGaussian, and incorporates symmetry priors to enhance the geometry and texture fidelity of single-image 3D generation.

\noindent\textbf{Symmetry for 3D Reconstruction.} Symmetry has long been recognized as a powerful cue for recovering 3D structures of objects~\cite{YANG2005210}. Recent works~\cite{wu2020unsup3d,wu2021derender,xu2020ladybird, yao2020front2back} study the utility of symmetry in the less constrained single-image 3D reconstruction problem. However, these methods are often constrained to specific object categories, such as faces~\cite{wu2020unsup3d} or rotational artifacts~\cite{wu2021derender}, or limited to small datasets covering a few categories~\cite{xu2020ladybird, yao2020front2back}. In contrast, our work builds on the insight of leveraging symmetry prior to aid 3D reconstruction in a more generalizable manner. We introduce a zero-shot symmetry detection approach extending beyond category-specific constraints. Unlike~\cite{xu2020ladybird, yao2020front2back} that can detect symmetry and reconstruct 3D shapes on about 15 categories, we train our symmetry detector on thousands of categories to enable generalization to in-the-wild imagery. Furthermore, we demonstrate the benefit of integrating symmetry cues in modern single-image 3D generation pipelines~\cite{tang2023dreamgaussian} guided by diffusion models. Using our accurate symmetry estimation, we achieve superior fidelity in the generated 3D structure and texture.




%% file: sec/3_method.tex

\section{Method}

In this section, we describe our single-image 3D symmetry detector, \methodname, and our symmetry-aware 3D generation approach. We begin by formalizing the symmetry detection problem in \Cref{sec:problem}. We then introduce our feed-forward symmetry detector in \Cref{sec:ff}. We discuss our multi-view symmetry enhancements pipeline in \Cref{sec:mv}. Finally, in \Cref{sec:sds}, we describe how we integrate our detected symmetry into SDS for single-image 3D generation.



\subsection{Problem Definition}\label{sec:problem}

\noindent\textbf{Reflection Symmetry.} We follow the definition of reflection symmetry in~\cite{zhou2021nerd}. Consider the set of points in homogeneous coordinates on the surface of a 3D shape, $S \subset \mathbb{R}^4$. The shape exhibits reflection symmetry with respect to a plane $p$, if there exists a plane $p$ such that for every point $\mathbf{x} \in S$, its reflected point $\mathbf{M}_p\mathbf{x}$ is also contained within $S$. Here, $\mathbf{M}_p$ is the reflection transformation matrix across plane $p$. The surface properties $F(\mathbf{x})$ at point $\mathbf{x}$ are preserved under reflection. Formally:

\vspace{-3ex}

\begin{align}
S = M_p(S)
\quad \text{and} \quad F(\mathbf{x}) = F(\mathbf{M}_p\mathbf{x})
\label{eq:symm}
\end{align}

\vspace{-1ex}

The reflection matrix $\mathbf{M}_p \in \mathbb{R}^{4 \times 4}$ corresponding to plane $p$ is defined as:

\vspace{-2ex}

\begin{align}
\mathbf{M}_p = \begin{bmatrix} \mathbf{I} - 2 \mathbf{n}_p \mathbf{n}_p^T & -2 d_p \mathbf{n}_p \\ 0 & 1 \end{bmatrix}
\end{align}

Where the plane $p$ is parameterized by its normal vector $\mathbf{n}_p \in \mathbb{R}^3$ and its distance $d_p$ from the origin. 

In~\Cref{eq:symm}, $M_p(S) = \{ \mathbf{M}_p\mathbf{x} \mid \mathbf{x} \in S \}$ represents the set of all points in $S$ reflected across the plane $p$, and $F(\mathbf{x})$ represents the surface properties at point $\mathbf{x}$, such as texture.

\noindent\textbf{Symmetry Detection.} In this work, we address the problem of single-image reflection symmetry detection. Given an object-centric image $I \in \mathbb{R}^{h \times w \times 3}$ with underlying 3D shape $S \subset \mathbb{R}^3$, we aim to find a set of symmetry planes $\mathcal{P}$, defined as:

\vspace{-3ex}
\begin{align}
\mathcal{P} = \{ p \mid p = (\mathbf{n}_p, d_p), \mathbf{n}_p \in \mathbb{R}^3, d_p \in \mathbb{R}, \|\mathbf{n}_p\| = 1 \}
\end{align}

Such that for each $p \in \mathcal{P}$, $S = M_p(S)$ and $F(\mathbf{x}) = F(\mathbf{M_px})$. $\mathcal{P}$ could be an empty set for completely asymmetric objects.



Due to single-view scale ambiguity, it is challenging to determine the absolute distance $d_p$ from the symmetry plane to the origin during symmetry detection. We follow~\cite{zhou2021nerd} to detect $\mathbf{n}_p \in \mathbb{R}^3$ and resolve the relative $d_p$ using other cues, \eg, make it consistent with generated 3D representation in symmetry-aware 3D generation.

\subsection{Feed-Forward Symmetry Detector} \label{sec:ff}

Inspired by the success of foundation models, we propose our feed-forward single-image symmetry detector \methodname, illustrated in the top section of~\Cref{fig:symm_method}.

\noindent\textbf{Image Encoder.} Previous works in correspondence learning~\cite{zhang2023tale, barroso2024mickey} have identified that DINOv2~\cite{dinov2} is sensitive to 3D spatial features of objects. 
In contrast to other foundation models, such as CLIP~\cite{clip}, which focuses more on object semantics, DINOv2 aligns better with our goal of detecting geometric cues that facilitate accurate symmetry prediction. We use a frozen DINOv2 encoder to encode our images into a set of features.

\noindent\textbf{Symmetry Decoder.} Given the spatial features from our encoder, we design a decoder that transforms these features into symmetry predictions. We discretize the space of possible symmetry plane normals by evenly sampling $N$ fixed unit vectors spanning a hemisphere, which serve as \textit{symmetry hypotheses}. Each hypothesis represents a small neighborhood containing all normal vectors closest to this normal. Hence our hypotheses cover all possible normals. For each hypothesis, we classify whether its neighborhood contains a ground-truth symmetry plane and regress the exact normal vector from the hypothesis. Note that our symmetry hypotheses only span a hemisphere rather than the entire unit sphere because a normal vector $\mathbf{n}$ and its opposite $\mathbf{-n}$ represent the same plane.


Our symmetry hypotheses are $N$ 3-dimensional vectors. We transform the $N$ hypotheses into high-dimensional query features with a shallow MLP. We then apply cross-attention between the query features and image features, followed by a set of self-attention and MLP layers.
This yields $N$ feature vectors, each corresponding to a symmetry hypothesis. For each feature vector, we apply an MLP head to perform binary classification of symmetry, and another MLP head to regress the rotation residual parameterized by a quaternion. Symmetry predictions are obtained by applying the quaternion to the corresponding positive hypothesis. Our detector also provides a confidence score for each prediction derived from the classification probability.


\noindent\textbf{Training.} To train our feed-forward symmetry detector, we need to assign each ground truth symmetry plane normal to a corresponding symmetry hypothesis.
We sample a sufficient number of hypotheses to ensure that the geodesic distance between each two ground truths is much larger than the geodesic distance between two hypotheses. Then each ground truth is matched to its nearest symmetry hypothesis. We supervise the symmetry binary classification using binary cross-entropy (BCE) loss, and supervise the quaternion regression using mean squared error (MSE) between the predicted normal and the ground truth. 

\begin{figure*}[t]
\vspace{-2ex}
  \centering
   \includegraphics[width=\textwidth,trim={3.5cm 7.3cm 3cm 7.5cm}, clip]{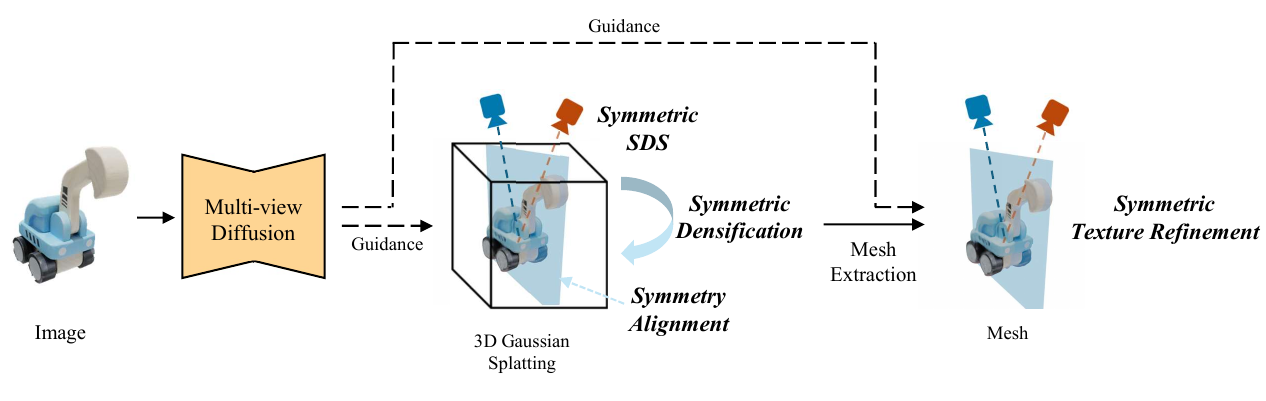}
   \vspace{-5ex}
   \caption{Our symmetry-aware 3D generation pipeline (\Cref{sec:sds}). Building on DreamGaussian~\cite{tang2023dreamgaussian}, we integrate the detected symmetry through three steps, namely, symmetry alignment, symmetric SDS optimization, and symmetric texture refinement.}
   \vspace{-2.5ex}
   \label{fig:shape_method}
\end{figure*}

\subsection{Multi-view Symmetry Enhancement}  \label{sec:mv}
A major difficulty in single-image symmetry prediction is the single-view ambiguity---the high uncertainty of the back surface of objects makes the learning ambiguous, especially under a regression framework. We explore using a multi-view diffusion model to resolve this ambiguity and enhance the robustness of our symmetry prediction (see ~\Cref{fig:symm_method} bottom).

To aggregate symmetry predictions from different views, we first apply the multi-view diffusion model conditioned on the input image, and generate $M$ views surrounding the origin. We then filter out inconsistent output images via CLIP-Similarity~\cite{clip}. We apply our feed-forward symmetry model to the remaining views to obtain a set of symmetry predictions. After rotating all predictions into the coordinate system of the input image view, we perform a K-Means clustering to aggregate the predictions. Finally, we take the cluster centers as our symmetry normal prediction.

\subsection{Single-image 3D Generation}\label{sec:sds}

We present a novel pipeline for generating 3D textured meshes from a single image, guided by our detected symmetry. Our method is illustrated in~\Cref{fig:shape_method}.

\noindent\textbf{Preliminary.} Our work is built on DreamGaussian. DreamGaussian addresses the single-image 3D generation through two stages: SDS optimization and UV texture refinement. 
In the SDS optimization stage, the scene is initialized as Gaussian splats distributed on a unit sphere. The optimization leverages two loss functions: an MSE loss to align with the input view image and an SDS loss, guided by a diffusion model, applied to randomly sampled novel views. After this stage, a mesh is extracted from the Gaussian splatting scene using the marching cubes algorithm~\cite{mcubes}. The texture is then refined in the UV space. The optimization target of the refinement includes an MSE loss applied to the input view and a texture refinement loss. To compute the texture refinement loss, we first add noise to the rendered images from random viewpoints. A 2D diffusion model is then applied to denoise these images. Finally, we calculate the MSE loss between the rendered and denoised images. We incorporate symmetry into DreamGaussian~\cite{tang2023dreamgaussian} in three stages: symmetry alignment, symmetric SDS optimization, and symmetric texture refinement.

\begin{table}[t]
  \centering
  \resizebox{0.48\textwidth}{!}{
  \begin{tabular}{lcccccc}
             & Base Dataset & \#Objects & \#Categories & \#Images &   \begin{tabular}{@{}c@{}}\#Symmetry \\ Planes \end{tabular}  \\
    \midrule
SN-ShapeNet~\cite{shi_siga20}& ShapeNet & 30,000 &11&300,000& 38,592$^*$ \\
SN-YCB~\cite{shi_siga20} & YCB &18 & 18&16,189&57$^*$ \\
NeRD-ShapeNet~\cite{zhou2021nerd} & ShapeNet & 35,030 & 13 & 175,122 & 35,030 \\
NeRD-Pix3D~\cite{zhou2021nerd} & Pix3D & 5,285 & 9 & 5,285 & 5,285 \\
\midrule
Ours         & \begin{tabular}{@{}c@{}}Objaverse \\ + ShapeNet\end{tabular} &84,789&1,154&1,095,949& 152,019\\
  \end{tabular}
  }
  \vspace{-2ex}
  \caption{Statistics of our curated dataset. Compared to datasets used in prior works, our curated data enjoys a much higher object diversity and image quantity. $^*$ denotes an estimate because the exact training datasets are not released in full. }
  \label{tab:data}
\end{table}


\noindent\textbf{Symmetry Alignment.} As described in~\Cref{sec:problem}, our symmetry detection process identifies the symmetry plane direction $\mathbf{n}_p \in \mathbb{R}^3$. In this stage, we optimize the 3D location and orientation of the symmetry plane to align it with the Gaussian scene. Specifically, we perform the first stage of DreamGaussian optimization without the MSE loss for a limited number of steps. This produces a coarse Gaussian representation with blurred texture and limited detail. From the coarse Gaussian, a point cloud is extracted, and the detected symmetry plane is aligned with the point cloud.  Specifically, we reflect the point cloud using the current symmetry estimation, and perform Iterative Closest Points (ICP) between the original and reflected point clouds. We provide detailed description of this alignment process in the appendix.

\noindent\textbf{Symmetric SDS Optimization.} In this stage, symmetry information is incorporated in two key steps. First, instead of sampling a single camera view for the SDS loss, we compute SDS losses for both the sampled view and its symmetric counterpart. Second, we symmetrically densify the Gaussian splatting periodically. Every 100 iterations, all Gaussians are reflected across the symmetry planes, and 50\% of these reflected Gaussians are randomly sampled and appended to the original set. These strategies encourage symmetry while allowing for minor deviations, accommodating the natural asymmetries often observed in real-world objects.

\noindent\textbf{Symmetric Texture Refinement.} In this stage, we extract a mesh from the 3D Gaussians and optimize the UV texture conditioned on the symmetry. Specifically, for mesh regions visible from the input view, the texture is refined directly using an MSE loss with the input image. For regions visible only from the mirrored view, we refine the UV texture using an MSE loss with a flipped image. The same texture refinement loss as DreamGaussian is applied to the remaining regions. Overall, this three-stage approach ensures that the generated 3D shape respects its inherent symmetry, enhancing consistency between the object's front and back sides.

\subsection{Implementation Details} \label{sec:impl_details}

We train our feed-forward symmetry detector using the Adam optimizer with a learning rate of $3e^{-5}$. We use a batch size of 120 and train for 15 epochs. We set the number of symmetry hypotheses $N$ to 31. Since reflection symmetry planes are often far apart in direction, we find this number of hypotheses is sufficient to prevent multiple symmetry planes from being assigned to the same query. We empirically discover that varying the number of hypotheses in a reasonable range has minimal impact on performance.
In multi-view enhancement, we set the number of novel views $M$ to 8. We found that 8 views are sufficient and the performance saturates with more views. More implementation details are available in the appendix.


%% file: sec/4_data.tex
\begin{table*}[t]
  \centering
  \vspace{-1ex}
  \resizebox{0.95\textwidth}{!}{
  \begin{tabular}{lccccc|ccccc}
  & \multicolumn{5}{c|}{GSO} & \multicolumn{5}{c}{OmniObject} \\
    \midrule
  Method           & F@5$^\circ$$\uparrow$   & F@15$^\circ$$\uparrow$  & F@30$^\circ$$\uparrow$  & F@50$^\circ$$\uparrow$  & GD$\downarrow$ & F@5$^\circ$$\uparrow$   & F@15$^\circ$$\uparrow$  & F@30$^\circ$$\uparrow$  & F@50$^\circ$$\uparrow$  & GD$\downarrow$ \\
    \midrule
Random Guess       & 0.007 & 0.043 & 0.190 & 0.464 & 48.4 & 0.007 & 0.062 & 0.187 & 0.374 & 52.1  \\
Shape to Symmetry & 0.021 & 0.128 & 0.283 & 0.785 & 34.9  & 0.023 & 0.112 & 0.236 & 0.587 & 34.6 \\
NeRD~\cite{zhou2021nerd}    & 0.040 & 0.398 & 0.461 & 0.564 & 36.2 & 0.055 & 0.166 & 0.328 & 0.590 & 41.3  \\
\midrule
\methodname-FF (Ours)  &  \underline{0.191} & \underline{0.452} & \underline{0.676} & \underline{0.867} & \underline{22.7} & \underline{0.103} & \underline{0.276} & \underline{0.478} & \underline{0.710} & \underline{31.1} \\
\methodname (Ours)    &  \textbf{0.390} & \textbf{0.756} & \textbf{0.848} & \textbf{0.889} & \textbf{13.3} & \textbf{0.173} & \textbf{0.474} & \textbf{0.650} & \textbf{0.799} & \textbf{22.8} \\
  \end{tabular}
  }
  \vspace{-1ex}
  \caption{Quantitative results of our symmetry detection method. Best results are in \textbf{bold}, and second-best results are \underline{underlined}. GD represents average geodesic distance. Our feed-forward model Reflect3D-FF, directly applied to the input image, already achieves state-of-the-art performance. Our multi-view enhanced \methodname delivers significant additional improvements.}
  \label{tab:symm_result}
  \vspace{-0.5ex}
\end{table*}

\begin{table*}[t]
  \centering
  \resizebox{0.95\textwidth}{!}{
  \begin{tabular}{lccccc|ccccc}
  & \multicolumn{5}{c|}{GSO} & \multicolumn{5}{c}{OmniObject} \\
    \midrule
  Method           & CLIP-Sim$\uparrow$ & CD$\downarrow$ &   F@0.1$\uparrow$  & F@0.2$\uparrow$  & F@0.5$\uparrow$  & CLIP-Sim$\uparrow$ & CD$\downarrow$ &  F@0.1$\uparrow$  & F@0.2$\uparrow$  & F@0.5$\uparrow$\\
    \midrule
DreamGaussian \cite{tang2023dreamgaussian} & 0.592 & 0.442 & 0.158 & 0.404 & 0.767 & 0.704 & 0.441 &  0.199 & 0.427 & 0.724\\
Ours          & \textbf{0.629}   & \textbf{0.414} & \textbf{0.172} & \textbf{0.451} & \textbf{0.827} & \textbf{0.734} & \textbf{0.421}  & \textbf{0.208} & \textbf{0.457} & \textbf{0.765}\\ 
  \end{tabular}
  }
  \vspace{-1ex}
  \caption{Quantitative results of our single-image 3D generation method for symmetric objects. CD represents Chamfer distance. Our approach improves both 2D and 3D fidelity compared to the baseline.}
  \label{tab:shape_result}
\end{table*}

\begin{table}[t]
  \centering
  \vspace{-1ex}
  \resizebox{0.48\textwidth}{!}{
  \begin{tabular}{lccccc}
  Method           & F@5$^\circ$$\uparrow$   & F@15$^\circ$$\uparrow$  & F@30$^\circ$$\uparrow$  & F@50$^\circ$$\uparrow$  & GD$\downarrow$ \\
    \midrule
\methodname (Ours)  &  0.390 & 0.756 & 0.848 & 0.889 & 13.3 \\
~~~~w/o clustering & 0.312 & 0.655 & 0.752 & 0.789 & 16.0\\
\midrule
\methodname-FF (Ours) & 0.191 & 0.452 & 0.676 & 0.867 & 22.7   \\
~~~~DINOv2 $\rightarrow$ ViT & 0.094 &  0.386 & 0.616 & 0.728 & 24.7 \\
~~~~Frozen $\rightarrow$ Finetune & 0.038 & 0.236 & 0.435 & 0.644 & 34.2 \\
  \end{tabular}
  }
  \vspace{-1.5ex}
  \caption{Ablation studies for our symmetry detector \methodname on the GSO dataset. Both the clustering-based multi-view aggregation and our frozen, geometric feature-aware encoder are crucial to symmetry detection performance.}
  \label{tab:symm_abl}
  \vspace{-2ex}
\end{table}



\section{Data Curation}
\label{sec:data}

\noindent\textbf{Training Data.} To scale up the training of our \methodname, we construct a large-scale symmetry detection dataset by combining Objaverse~\cite{objaverse} and ShapeNet~\cite{chang2015shapenet}, resulting in a diverse and extensive 3D reflection symmetry dataset. For the Objaverse dataset, we utilize only the high-quality subset corresponding to LVIS~\cite{gupta2019lvis} categories. Symmetry ground truth is generated using an optimization-based approach.
Specifically, we uniformly sample many candidate planes, reflect the 3D shape by all the candidate planes, and use Chamfer distance to verify whether this candidate likely corresponds to a symmetry plane, followed by an ICP-based refinement for the detected symmetry plane. More details are available in the appendix. The generated symmetry ground truth has been verified through manual inspection of a random subset to ensure alignment with human perception.
~\Cref{tab:data} highlights the advantages of our dataset compared to previous symmetry datasets. Our dataset offers significantly greater diversity in object categories and more comprehensive coverage of symmetry planes, as reflected in the higher number of symmetry planes per object.

\noindent\textbf{Evaluation Data.} Previous works on symmetry detection, such as~\cite{shi_siga20, zhou2021nerd, lin2021nerd++}, evaluate their methods on a validation split of their training datasets. In contrast, we evaluate our symmetry detection model, \methodname, in a more challenging approach to assess generalization to novel, unseen objects. Specifically, we evaluate on two scanned real-world 3D object datasets: Google Scanned Object (GSO)~\cite{gso} and OmniObject~\cite{omniobj}. The ground truth for these test datasets is generated and validated using the same methodology employed for the training data. We use all symmetric objects from GSO (572 objects) and a randomly sampled subset of OmniObject3D (100 objects). We also test our single-image 3D generation methods on the same test sets. These test sets provide a comprehensive assessment of models' generalization ability across diverse semantic categories and domains.

%% file: sec/5_experiments.tex
\begin{figure}[t]
  \centering
    \vspace{-2ex}
   \includegraphics[width=0.5\textwidth,trim={3.5cm 6cm 2cm 6.4cm}, clip]{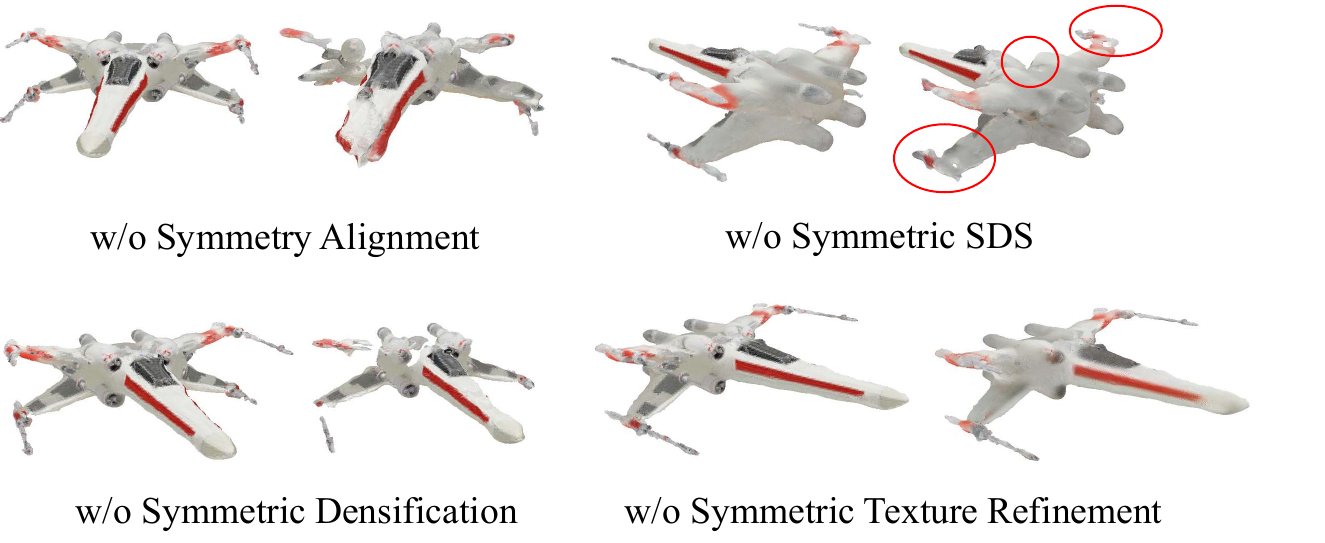}
   \vspace{-3.5ex}
   \caption{Ablation studies for our single-image 3D generation pipeline. Removing each component adversely affects geometry quality, texture quality, or both.}
   \label{fig:shape_abl}
   \vspace{-1.5ex}
\end{figure}

\section{Symmetry Detection Results}

\noindent\textbf{Evaluations Metrics.} 
Given a set of predicted symmetry plane normals $U = \{\mathbf{u_1}, \mathbf{u_2}, ...\}$ and ground truth normals $V = \{\mathbf{v_1}, \mathbf{v_2}, ...\}$, we evaluate predictions $U$ against ground truth $V$ with two metrics based on geodesic distance. The geodesic distance between any two normals $\mathbf{u}$ and $\mathbf{v}$, defined as $\theta(\mathbf{u}, \mathbf{v}) = \cos^{-1}(\mathbf{u} \cdot \mathbf{v})$, measures the angular difference between them.

We calculate F-scores at various geodesic distance thresholds. Given a threshold $\phi$ in degrees, a predicted and ground truth normal pair is considered a match if their geodesic distance is below $\phi$. The F@$\phi$ metric is the harmonic mean of precision@$\phi$ and recall@$\phi$. Intuitively, F@$\phi$ represents the percentage of correct predictions within a specified threshold.

We also calculate an average geodesic distance (GD). This metric evaluates both the \textit{exactness} and \textit{completeness} of predictions, similar to the precision-recall trade-off in F-scores. For exactness, each predicted normal $\mathbf{u}$ is matched to its closest ground truth normal $\mathbf{v^*}$, and their average geodesic distance $\theta_p$ is calculated. For completeness, each ground truth normal $\mathbf{v}$ is matched to its closest predicted normal $\mathbf{u^*})$, yielding an average geodesic distance $\theta_r$. The final metric is computed as $\theta_{avg} = (\theta_p + \theta_r)/2$.


\noindent\textbf{Baseline: Shape to Symmetry.} 
A straightforward baseline for single-image 3D symmetry detection is to first generate a 3D shape conditioned on the image, and then detect symmetry planes from the 3D shape. We use our single-image 3D generation baseline~\cite{tang2023dreamgaussian} to generate 3D shapes from input images. Then we apply our method for generating symmetry ground truth to these 3D shapes. The details for this method are discussed in~\Cref{sec:data} and the appendix.

\noindent\textbf{Baseline: NeRD~\cite{zhou2021nerd}.} NeRD represents the state-of-the-art in single-image 3D reflection symmetry detection. 
NeRD iteratively samples candidate symmetry plane normals following a coarse-to-fine strategy. Each candidate normal, together with known camera intrinsics, is used to warp 2D image features and construct a 3D cost volume. The cost volume is processed by 3D convolutions to compute a symmetry score. The candidate with the highest score is selected as the final prediction. We use the official implementation of NeRD and report the performance of the best-performing checkpoint on our test sets.





\noindent\textbf{Symmetry Results.} 
The results of our symmetry detection experiments are summarized in~\Cref{tab:symm_result}. Even without any multi-view information, our feed-forward model already significantly outperforms all baselines. Specifically, our feed-forward method is both 2 to 4 times more precise than prior state-of-the-art methods reflected in F@5$^\circ$, and also more comprehensive than previous methods as shown in F@50$^\circ$. Incorporating multi-view information further enhances performance, leading to a 1.7 to 2 times improvement in F@5$^\circ$ and reducing the average geodesic distance by 9.4$^\circ$ and 8.3$^\circ$ on the two datasets respectively. These results validate the effectiveness of scaling up our feed-forward transformer model and highlight the substantial gains from leveraging multi-view information to mitigate single-view ambiguity.

\begin{figure}[t]
\vspace{-1ex}
  \centering
   \includegraphics[width=0.5\textwidth,trim={2cm 1cm 0.5cm 1cm}, clip]{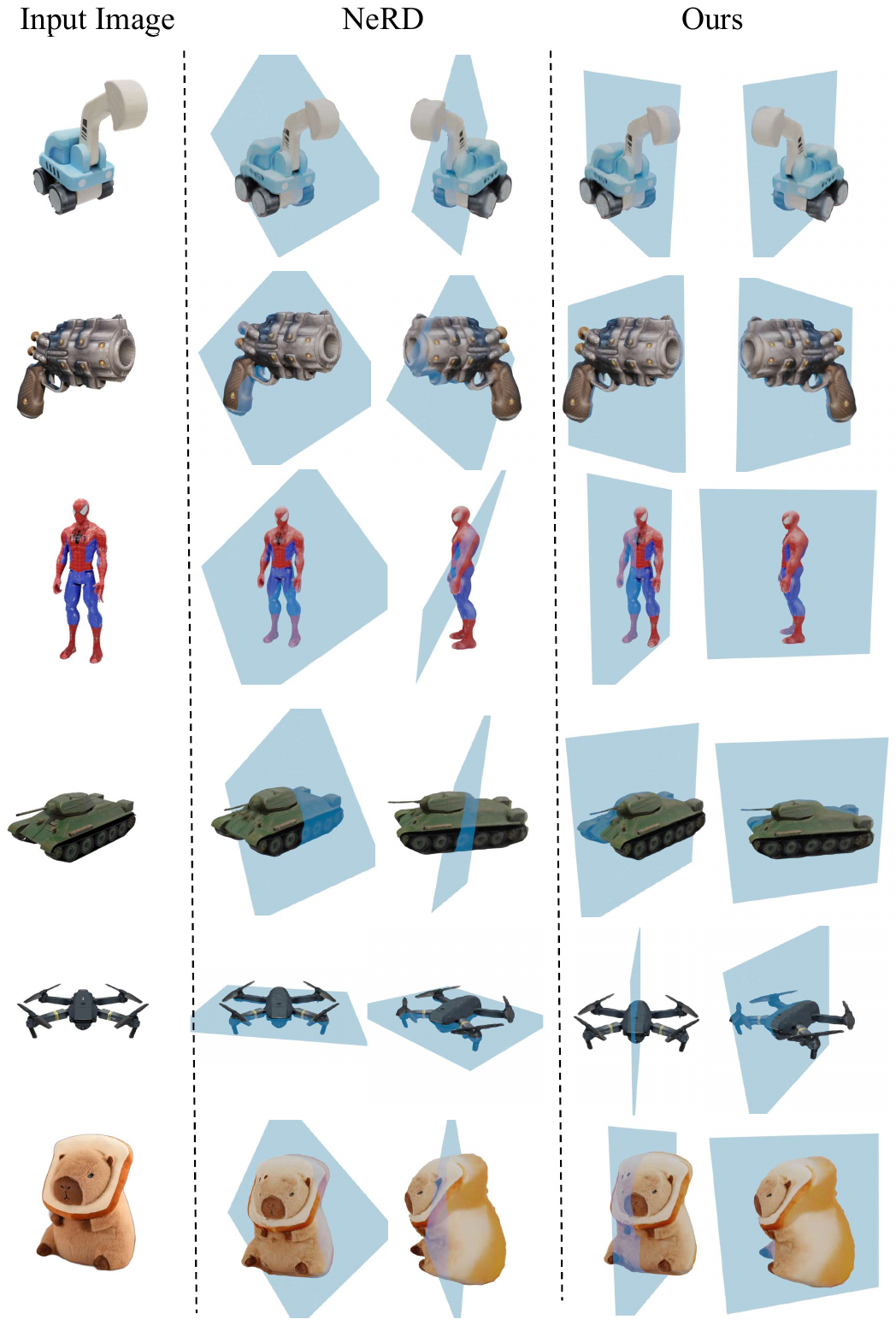}
   \vspace{-3ex}
   \caption{Qualitative results for our symmetry detection pipeline. Our \methodname achieves better generalization and precision than NeRD~\cite{zhou2021nerd}.  Please refer to our project webpage for video results.}
   \vspace{-1ex}
   \label{fig:symm_qual}
\end{figure}

Qualitative results in~\Cref{fig:symm_qual} further demonstrate the robustness of our approach. NeRD struggles to generalize to unseen novel objects, often incorrectly predicting a false front-to-back symmetry plane. In contrast, \methodname achieves accurate symmetry predictions for these complex, unseen objects.



\noindent\textbf{Symmetry Ablations.} We perform ablation studies on the GSO dataset to validate our design decisions in \methodname. 
In our multi-view enhancement pipeline, clustering the predictions from different views is crucial, since this eliminates redundant predictions that are close together and correspond to the same underlying symmetry plane. In our feed-forward model, the choice of feature representation is crucial. Replacing DINOv2 with ImageNet-trained ViT leads to less precise and less comprehensive predictions. Freezing DINOv2 and only training the decoder is also essential, as fine-tuning drastically degrades the performance. We speculate the pre-trained DINOv2 feature is key to the generalizability of our approach.

\section{Single-Image 3D Generation Results}

\noindent\textbf{Dataset and Metric.} We evaluate single-image 3D generation on the same datasets used for symmetry evaluation, \ie, GSO and OmniObject3D. 
The evaluation employs both 2D and 3D metrics. For 2D appearance quality, we calculate CLIP similarity following~\cite{tang2023dreamgaussian}. For 3D geometric quality, we calculate Chamfer distance and F-score following~\cite{huang2024zeroshape}. Specifically, for 3D evaluation, we first leverage ambient occlusion to remove the internal surface. Then we uniformly sample a large number of points from both the predicted and ground truth meshes to calculate Chamfer distance and F-score at various distance thresholds.


\noindent\textbf{Baseline: DreamGaussian~\cite{tang2023dreamgaussian}.} DreamGaussian introduces an efficient and high-quality framework for image-conditioned 3D content generation based on Score Distillation Sampling (SDS). Details of the DreamGaussian pipeline are introduced in~\Cref{sec:sds}. 
We select the best-performing image-conditioned variant of DreamGaussian, which uses Stable-Zero123~\cite{stablezero123} as guidance in the SDS optimization. 

\begin{figure}[t]
\vspace{-1ex}
  \centering
    \includegraphics[width=0.5\textwidth,trim={2.2cm 2cm 1.7cm 2cm}, clip]{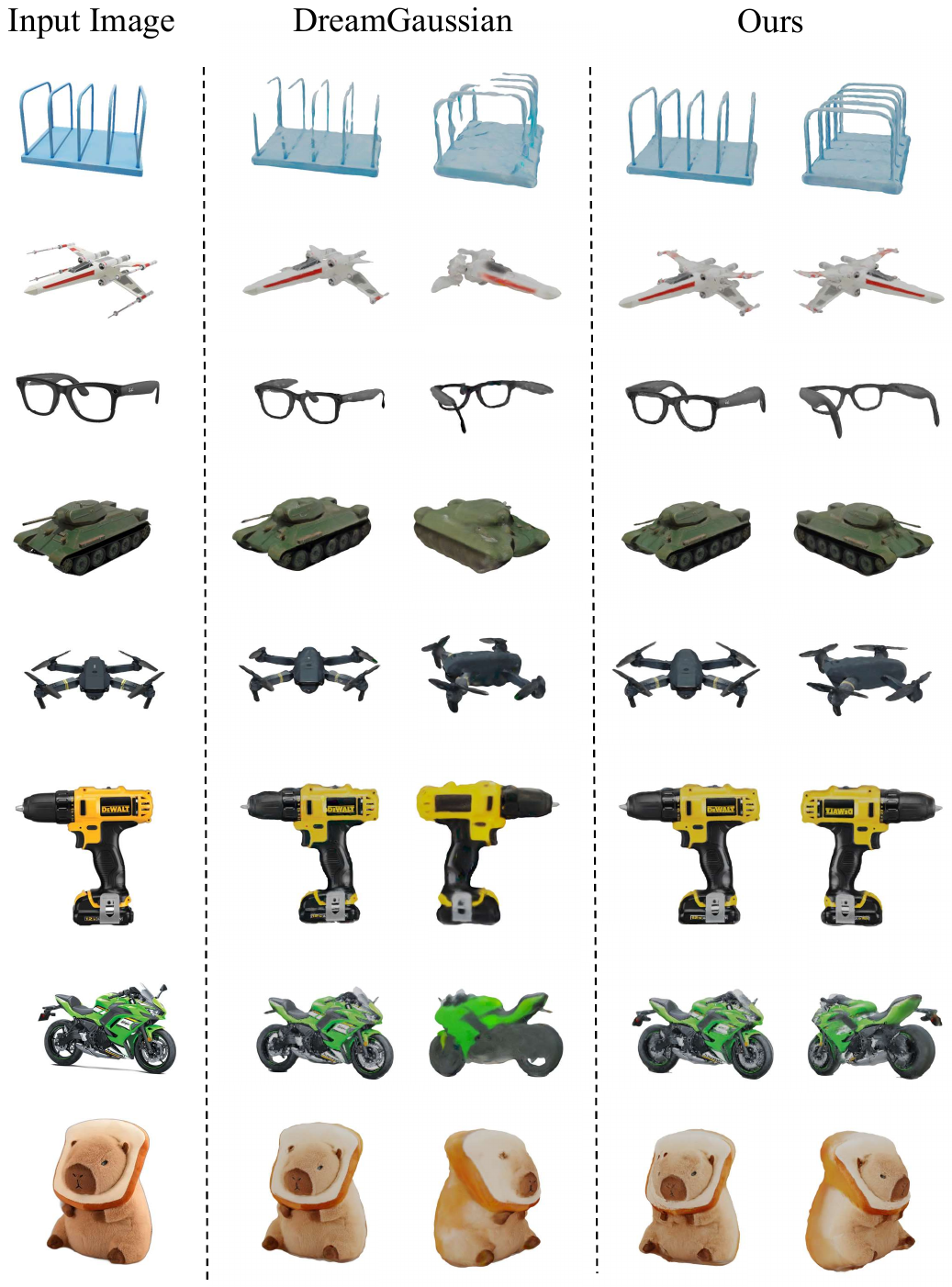}
    \vspace{-4ex}
   \caption{Qualitative results for our symmetry-conditioned single-image 3D method. Leveraging detected symmetry, our method avoids missing details and corrects geometric errors (top 4 rows). Meanwhile, we greatly improve the texture and geometric quality of the unobserved object backside (bottom 3 rows). Please refer to our project webpage for video results.}
   \label{fig:shape_qual}
\end{figure}

\noindent\textbf{3D Generation Results.} We summarize our symmetry-conditioned single-image 3D generation results in~\Cref{tab:shape_result}. We show that by integrating the symmetry prior, we effectively improve 3D generation in both 2D and 3D metrics. Qualitative comparisons in~\Cref{fig:shape_qual} highlight the benefit of incorporating the symmetry prior. Symmetry helps to avoid missing details (rows 1, 2, and 5) and structural errors. For example, in the baseline results, the left leg of the glasses (row 3) is misinterpreted as part of the circular frame, and the tank barrel (row 4) is incorrectly connected to the base rather than the turret. Our method rectifies these issues. Additionally, the unobserved back side of the baseline-generated objects often lacks geometric detail and texture, whereas our approach leverages symmetry to produce significantly improved geometry and textures for the back side (rows 6, 7, and 8).

\noindent\textbf{Single-Image 3D Generation Ablations.} 
Ablation studies for our method are shown in~\Cref{fig:shape_abl}. Each of the four components in our pipeline is critical to achieving high-quality geometry and texture. Removing symmetry alignment results in inaccurate symmetry, causing distorted appearances. Removing symmetrically sampled SDS leads to lower overall quality and worse symmetry in details, as highlighted in red. Omitting symmetric densification produces sparse outputs that fail to fully respect symmetry. Finally, removing symmetric texture refinement results in blurry textures on the unobserved back side. The synergy of these components is essential for the superior geometry and texture quality achieved by our method.



%% file: sec/6_conclusion.tex
\section{Limitations}

A majority of man-made objects and a significant proportion of natural objects are symmetrical. However, there exist objects that are completely asymmetric or highly deformable. Our approach is not designed to handle these objects. However, as mentioned in~\Cref{sec:sds}, our method does have tolerance for asymmetric details.
Addressing the challenges for these objects remains an avenue for future work.

\section{Conclusion}

We propose \methodname, a scalable zero-shot approach for single-image reflection symmetry detection. Through improvements in architecture, data, and our multi-view aggregation to tackle single-view ambiguity, we significantly improve the generalization performance of single-image symmetry detection. Furthermore, we empirically demonstrate that our detected symmetry can enhance SDS-based single-image 3D generation pipelines, resulting in substantial improvements in the structural and textural quality of the generated objects.

%% file: sec/X_suppl.tex
\clearpage

\section{Appendix}

In this appendix, we present 1) a detailed description of our ground truth symmetry plane generation method; 2) more implementation details for our \methodname symmetry detector and our single-image 3D generation pipeline. All references and citations in this supplementary document refer to the main paper.




\subsection{Dataset}

In this section, we provide more details on generating ground truth reflection symmetry for our training and evaluation datasets. This generation method is also used in our \textit{Shape to Symmetry} baseline method.

\noindent\textbf{Automatic Ground Truth Generation.} We propose a reliable protocol to efficiently generate accurate symmetry ground truth for arbitrary 3D datasets. First, for each mesh in the 3D dataset, we center it by its bounding sphere center and normalize it by scaling its bounding sphere to a unit sphere. This ensures the center of any potential ground truth symmetry plane is on the origin. Then we uniformly sample $N_p$ points from this mesh. To find the symmetry plane, we generate a set of $N_c$ unit vectors uniformly spanning a unit hemisphere as candidates for symmetry plane normals, analogous to the symmetry hypothesis in our \methodname single-image symmetry detector. These candidates only span a hemisphere rather than the entire unit sphere because a normal vector $\mathbf{n}$ and its opposite $-\mathbf{n}$ represent the same plane. For each candidate normal, we derive its corresponding candidate plane as the plane passing through the origin and having a normal as the candidate normal. We choose a large enough number of sampled points $N_p$ 
 and symmetry candidates $N_c$, empirically we use $N_p = 50000$ and $N_c = 31$. 

We reflect the point cloud by each candidate plane. Then by how well the reflected point cloud and the original point cloud align, we can infer whether the plane is a ground truth symmetry plane of the shape. We calculate a Chamfer distance between the original and reflected point clouds. We manually select a threshold to eliminate low-quality symmetry planes. We determine this threshold after comprehensively examining the results and making sure it aligns with human perception of symmetry. Meanwhile, for candidate planes that pass the threshold, we can refine them by registering the reflected point cloud to the original point cloud. We apply the iterative closest point (ICP) algorithm for this registration. Then we derive a refined symmetry plane from the registered point clouds.

\subsection{Implementation Details}

\noindent\textbf{\methodname.} Our method employs a frozen DINOv2 backbone with the ViT-L/14 architecture. The symmetry decoder consists of 4 layers with 256 channels. Our Adam optimizer uses a weight decay of 0.05. The classification loss and regression loss are weighted at 1.0 and 0.1, respectively. Training is performed on 2 A100 GPUs for 2 days. For multi-view enhancement, we generate 8 views with the same elevation as the input view, uniformly spacing them in azimuth such that adjacent views are $45^\circ$ apart. Finally, a clustering threshold of $30^\circ$ is applied in the aggregation step.

\noindent\textbf{Single-Image 3D Generation.} During symmetry alignment, we disable the image-based MSE loss and apply only the symmetrically sampled SDS loss based on the detected symmetry. We run this optimization stage for 400 steps to obtain more accurate symmetry planes, though fewer steps often suffice in practice. We convert the rough Gaussian splatting to a mesh using the marching cubes algorithm. We remove internal surfaces via ambient occlusion, and uniformly sample 10,000 points from the mesh surface to align the symmetry plane, as discussed in~\Cref{sec:sds}. In the subsequent symmetric SDS optimization stage, we optimize for 500 steps following DreamGaussian~\cite{tang2023dreamgaussian}. Finally, we refine the texture for 50 steps.